%% file: main.tex
\documentclass[runningheads]{llncs}
\usepackage{amsmath,amssymb} % define this before the line numbering.

\input{setup/package}
\input{setup/macros}

\input{setup/symbols}

\input{setup/graphicspath}

\usepackage[width=122mm,left=12mm,paperwidth=146mm,height=193mm,top=12mm,paperheight=217mm]{geometry}
\begin{document}
\pagestyle{headings}
\mainmatter
\def\ECCVSubNumber{3007}  % Insert your submission number here

\title{Learning to Learn in a Semi-Supervised Fashion} % Replace with your title

\titlerunning{Learning to Learn in a Semi-Supervised Fashion}

\author{
Yun-Chun Chen\inst{1} \and
Chao-Te Chou\inst{1} \and
Yu-Chiang Frank Wang\inst{1,2}
}

\authorrunning{Y.-C. Chen, C.-T. Chou, and Y.-C. F. Wang}

\institute{
Graduate Institute of Communication Engineering, National Taiwan University, Taiwan \and
ASUS Intelligent Cloud Services, Taiwan \\
% \email{\{\url{b03901148}, \url{b03901096}, \url{ycwang}\}\url{@ntu.edu.tw}}
}

\maketitle

\input{abstract.tex}

\input{intro.tex}
\input{related_work.tex}
\input{method.tex}
\input{experiment.tex}
\input{conclusion.tex}

{\flushleft {\bf Acknowledgments.}} 
This paper is supported in part by the Ministry of Science and Technology (MOST) of Taiwan under grant MOST 109-2634-F-002-037.

\newpage
\bibliographystyle{splncs04}
\bibliography{reference}
\end{document}

%% file: setup/package.tex
\usepackage{graphicx}
\usepackage{subcaption}
\usepackage{float}
\usepackage{caption}	% makes captions ragged right - thanks to Bryce Lobdell
\usepackage{lscape}                                         % Useful for wide tables or figures.

\usepackage[lined,ruled,linesnumbered]{algorithm2e}

\usepackage{booktabs}                   % Publication quality tables
\usepackage{multirow}

\usepackage{paralist}
\usepackage{enumitem}

\usepackage{bm}                          % Make bold, italic math symbols
\usepackage{epsfig}                      % for figures
\usepackage{graphicx}                  % another package that works for figures
\usepackage{mathtools}

\usepackage{color}

\usepackage{comment}

\usepackage{url}  % Hyphenation of URLs.
\usepackage[pagebackref=false,breaklinks=true,colorlinks,urlcolor=blue,citecolor=blue,linkcolor=blue,bookmarks=false]{hyperref}
\usepackage[nocompress]{cite}

\usepackage{listings}

\usepackage{xspace}
\usepackage[table]{xcolor}
\usepackage{setspace}

\usepackage{nicefrac}
\usepackage{microtype}
\usepackage[utf8]{inputenc} % allow utf-8 input
\usepackage[T1]{fontenc}    % use 8-bit T1 fonts

%% file: setup/macros.tex
\newlength\paramargin
\newlength\figmargin
\newlength\tablemargin
\newlength\secmargin
\newlength\figcapmargin
\newlength\tablecapmargin
\newlength\itemizemargin

\newcommand{\ra}[1]{\renewcommand{\arraystretch}{#1}}

\long\def\ignorethis#1{}

\newcommand{\mpage}[2]
{
\begin{minipage}{#1\linewidth}\centering
#2
\end{minipage}
}

%% file: setup/symbols.tex
\def\xi{\mathbf{x}_i}

%% file: setup/graphicspath.tex
\graphicspath{{figure}, {images}, {example}}

%% file: abstract.tex
\begin{abstract}
To address semi-supervised learning from both labeled and unlabeled data, we present a novel meta-learning scheme. We particularly consider that labeled and unlabeled data share disjoint ground truth label sets, which can be seen tasks like in person re-identification or image retrieval. Our learning scheme exploits the idea of leveraging information from labeled to unlabeled data. Instead of fitting the associated class-wise similarity scores as most meta-learning algorithms do, we propose to derive semantics-oriented similarity representations from labeled data, and transfer such representation to unlabeled ones. Thus, our strategy can be viewed as a self-supervised learning scheme, which can be applied to fully supervised learning tasks for improved performance. Our experiments on various tasks and settings confirm the effectiveness of our proposed approach and its superiority over the state-of-the-art methods.
\end{abstract}

%% file: intro.tex
\section{Introduction}

Recent advances of deep learning models like convolutional neural networks (CNNs) have shown encouraging performance in various computer vision applications, including image retrieval~\cite{xu2018unsupervised,tieu2004boosting,swets1996using} and person re-identification (re-ID)~\cite{chen2019learning,li2019recover,farenzena2010person,zheng2015scalable,li2020cross}. Different from recognizing the input as a particular category, the above tasks aim at learning feature embeddings, making instances of the same type (e.g., object category) close to each other while separating those of distinct classes away. Similar tasks such as image-based item verification~\cite{liu2016deepfashion}, face verification~\cite{taigman2014deepface}, face recognition~\cite{deng2019arcface,schroff2015facenet,sun2014deep}, and vehicle re-ID~\cite{Chu_2019_ICCV,Tang_2019_ICCV,Wang_2019_ICCV} can all be viewed as the tasks of this category.

\begin{figure}[t]
  \begin{subfigure}[b]{0.49\linewidth}
    \centering
    \includegraphics[width=1\linewidth]{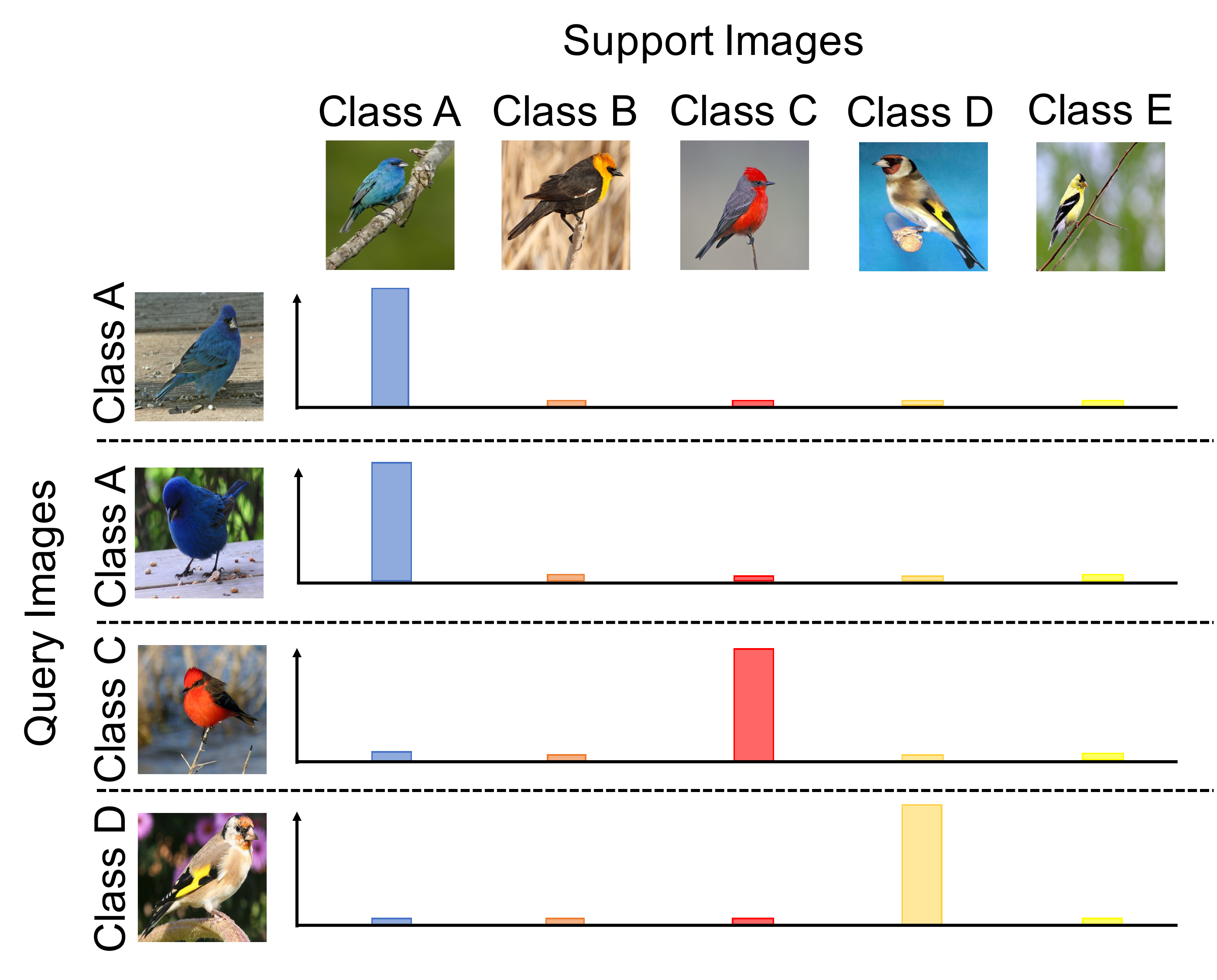}
    \caption{Existing meta-learning methods.}
    \label{fig:teaser-few-shot}
  \end{subfigure}
  \hfill
  \begin{subfigure}[b]{0.49\linewidth}
    \centering
    \includegraphics[width=1\linewidth]{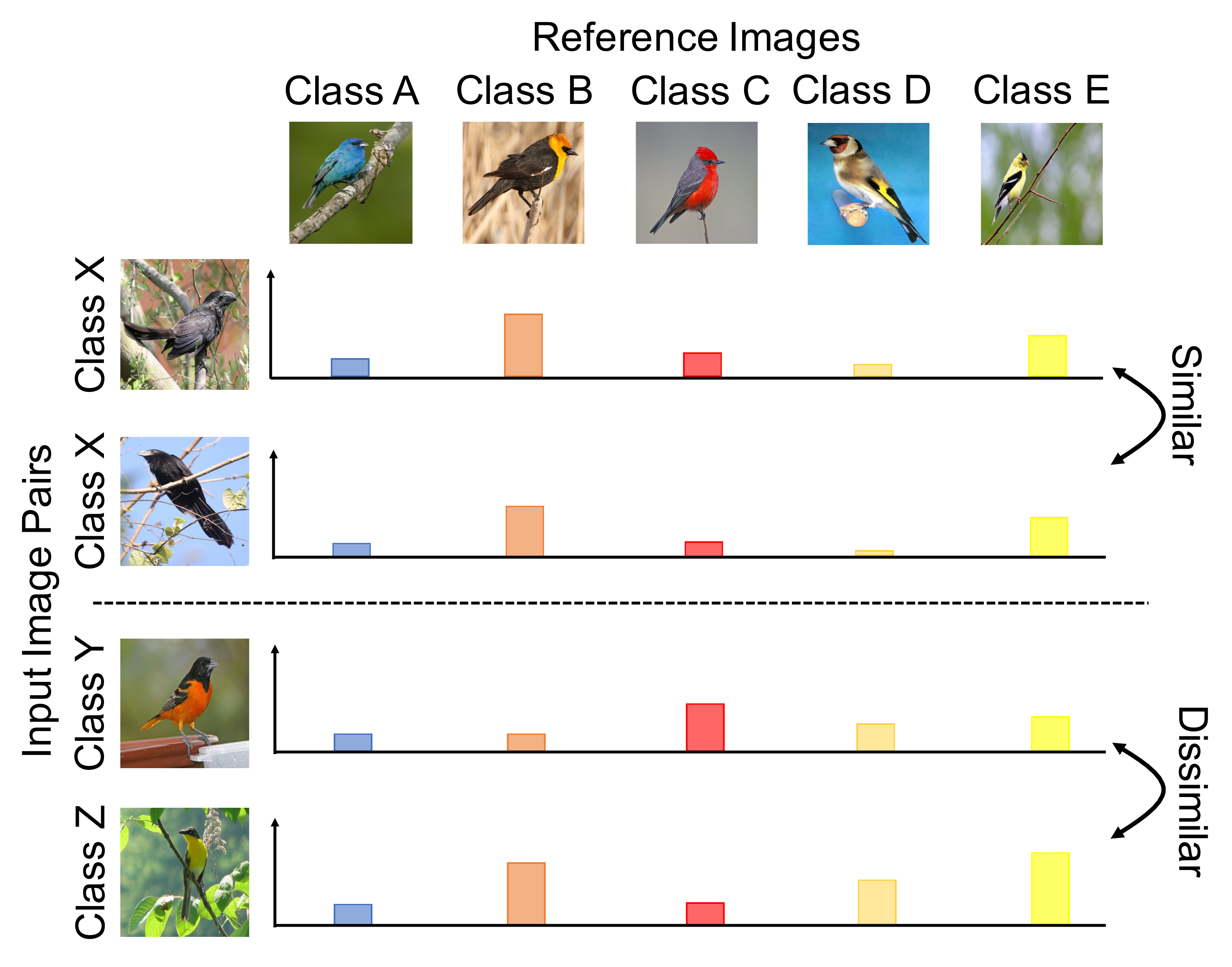}
    \caption{Our meta-learning algorithm.}
    \label{fig:teaser-ours}
  \end{subfigure}
  \caption{\textbf{Illustration of learning class-wise similarity.} (a) Standard meta-learning methods for visual classification compute class-wise similarity \textit{scores} between the query image and those in the support set, where the two sets share \textit{overlapping} ground truth labels. (b) To deal with training data with non-overlapping labels, our meta-learning scheme derives semantics-oriented similarity representations in a learning-to-learn fashion, allowing the determination of pairwise relationship between images with unseen labels.}
  \label{fig:teaser}
\end{figure}

Existing methods for image matching generally require the collection of a large number of labeled data, and tailor algorithms to address the associated tasks (e.g., image retrieval~\cite{tieu2004boosting,swets1996using} and person re-ID~\cite{zheng2016person,li2019recover,farenzena2010person,zheng2015scalable}). However, the assumption of having a sufficient amount of labeled data available during training may not be practical. To relax the dependency of manual supervision, several \textit{semi-supervised} methods for image retrieval~\cite{Wang_2018_ECCV,zhang2017ssdh,wang2012semi} and person re-ID~\cite{li2018semi,wu2019progressive} are proposed. These methods focus on learning models from datasets where each category is partially labeled (i.e., some data in \textit{each} category are labeled, while the rest in that category remain unlabeled). Thus, they choose to use the models learned from the labeled data to assign pseudo labels to the unlabeled ones~\cite{zhang2017ssdh,wang2012semi,wu2019progressive}, or adopt ensemble learning techniques to enforce the predictions of the unlabeled data to be consistent across multiple networks~\cite{Wang_2018_ECCV}. Despite significant progress having been reported, these methods \emph{cannot} be directly applied to scenarios where novel objects or persons are present.

To deal with instances of unseen categories for image matching purposes, one can approach such problem in two different ways. The majority of existing methods focuses on the cross-dataset (domain adaptation~\cite{hoffman2018cycle,chen2019pixel}) setting, where one dataset is fully labeled (i.e., source domain dataset) while the other one remains unlabeled. (i.e., target domain dataset)~\cite{yu2019unsupervised,deng2018image,wang2018transferable}. Existing methods for this category typically assume that there is a domain gap between the two datasets. These methods either leverage adversarial learning strategies to align feature distributions between the two datasets~\cite{deng2018image,wang2018transferable}, or aim at assigning pseudo labels for each unlabeled image in the target dataset through predicting class-wise similarity scores from models trained on the source (labeled) dataset~\cite{yu2019unsupervised}. By carefully selecting hyperparameters such as the prediction score threshold, one can determine whether or not a given image pair from the target dataset is of the same category. However, the class-wise similarity scores are computed based on a network trained on the source dataset, which might not generalize well to the target dataset, especially when their labels are non-overlapping. On the other hand, these methods are developed based on the assumption that a large-scale labeled dataset is available. 

Another line of research considers learning models from a \emph{single} dataset, in which only some categories are fully labeled while the remaining classes are unlabeled~\cite{xin2019semi,xin2019deep}. These methods typically require the number of classes of the unlabeled data to be known in advance, so that one can perform clustering-like algorithms with the exact number of clusters for pseudo label assignment. Having such prior knowledge, however, might not be practical for real-world applications.

In this paper, we propose a novel meta-learning algorithm for image matching in a semi-supervised setting, with applications to image retrieval and person re-ID. Specifically, we consider the same semi-supervised setting as \cite{xin2019semi,xin2019deep}, in which the ground truth label sets of labeled and unlabeled training data are \textit{disjoint}. Our meta-learning strategy aims at exploiting and leveraging class-wise similarity representation across labeled and unlabeled training data, while such similarity representation is derived by a learning-to-learn fashion. The resulting representations allow our model to relate images with pseudo labels in the unlabeled set (e.g., Figure~\ref{fig:teaser-ours}). This is very different from existing meta-learning for visual classification methods (like few-shot learning), which typically assume that the support and query sets share the same label set and focus on fitting the associated class-wise similarity scores (e.g., Figure~\ref{fig:teaser-few-shot}). Our learning scheme is realized by learning to match randomly selected labeled data pairs, and such concepts can be applied to observe both labeled and unlabeled data for completing the semi-supervised learning process.

The contributions of this paper are summarized as follows:
\begin{itemize}
  \item We propose a meta-learning algorithm for image matching in semi-supervised settings, where labeled and unlabeled data share non-overlapping categories.
  \item Our learning scheme aims at deriving semantics-oriented similarity representation across labeled and unlabeled sets. Since pseudo labels can be automatically assigned to the unlabeled training data, our approach can be viewed as a self-supervised learning strategy.
  \item With the derivation of semantics-oriented similarity representations, our learning scheme can be applied to fully supervised settings and further improves the performance.
  \item Evaluations on four datasets in different settings confirm that our method performs favorably against existing image retrieval and person re-ID approaches.
\end{itemize}

%% file: related_work.tex
\section{Related Work}

{\flushleft {\bf Semi-supervised learning.}}
Semi-supervised learning for visual analysis has been extensively studied in the literature. Most of the existing methods focus on image classification and can be categorized into two groups depending on the learning strategy: 1) labeling-based methods and 2) consistency-based approaches. Labeling-based methods focus on assigning labels to the unlabeled images through pseudo labeling~\cite{lee2013pseudo}, label propagation~\cite{luo2018smooth}, or leveraging regularization techniques for performing the above label assignment~\cite{grandvalet2005semi}. Consistency-based approaches, on the other hand, exploit the idea of cycle consistency~\cite{zhu2017unpaired,chen2018deep,chen2020show} and adopt ensemble learning algorithms to enforce the predictions of the unlabeled samples to be consistent across multiple models~\cite{athiwaratkun2018there,laine2016temporal,qiao2018deep,tarvainen2017mean,berthelot2019mixmatch,sajjadi2016regularization,miyato2018virtual}. In addition to image classification, another line of research focuses on utilizing annotation-free images to improve the performance of semantic segmentation~\cite{souly2017semi,hung2018adversarial}. These methods adopt generative adversarial networks (GANs)~\cite{goodfellow2014generative} to generate images conditioned on the class labels to enhance the learning of feature representations for the unlabeled images~\cite{souly2017semi}, or develop a fully convolutional discriminator to generate dense probability maps that indicate the confidence of correct segmentation for each pixel in the unlabeled images~\cite{hung2018adversarial}.

To match images of the same category, a number of methods for image retrieval~\cite{Wang_2018_ECCV,zhang2017ssdh,wang2012semi} and person re-ID~\cite{li2018semi,wu2019progressive,figueira2013semi,liu2014semi,huang2018multi,ding2019feature,liu2018transductive} also consider learning models in semi-supervised settings. Methods for semi-supervised image retrieval can be grouped into two categories depending on the adopted descriptors: 1) hand-crafted descriptor based methods and 2) approaches based on trainable descriptors. The former typically focuses on optimizing the errors on the labeled set and leverages a regularizer to maximize the information entropy between labeled and unlabeled sets~\cite{wang2012semi}. Trainable descriptor based approaches either utilize a graph to model the relationship between labeled and unlabeled sets~\cite{zhang2017ssdh} or leverage a GAN~\cite{goodfellow2014generative} to learn triplet-wise information from both labeled and unlabeled data~\cite{Wang_2018_ECCV}. Similarly, methods for semi-supervised person re-ID also aim at relating labeled and unlabeled images through dictionary learning~\cite{liu2014semi}, multi-feature learning~\cite{figueira2013semi}, pseudo labeling with regularizers~\cite{huang2018multi}, or considering complex relationships between labeled and unlabeled images~\cite{ding2019feature}. While promising performance has been shown, these methods cannot be directly applied to scenarios where datasets contain labeled and unlabeled images with non-overlapping category labels, which are practical in many real-world applications.

To tackle this issue, two recent methods for semi-supervised person re-ID are proposed~\cite{xin2019semi,xin2019deep}. These methods either combine K-means clustering and multi-view clustering~\cite{xin2019semi}, or develop a self-paced multi-view clustering algorithm~\cite{xin2019deep} to assign pseudo labels to images in the unlabeled set. However, these methods require the number of identities of the unlabeled set to be known in advance. Our work does not need such prior knowledge. As noted above, we approach such problems and assign pseudo labels to the unlabeled data by learning their semantics-oriented similarity representations, which are realized in a unique learning-to-learn fashion.

{\flushleft {\bf Meta-learning.}}
The primary objective of meta-learning is to enable a base learning algorithm which observes data with particular properties to adapt to similar tasks with new concepts of interest. Few-shot learning~\cite{Snell-2017-NIPS,Sung-2018-CVPR} and neural architecture search~\cite{nasdip} are among the popular applications of meta-learning. Existing meta-learning algorithms can be grouped into three categories based on the learning task: 1) initialization-based methods, 2) memory-based approaches, and 3) metric-based algorithms. Initialization-based methods focus on learning an optimizer~\cite{Andrychowicz-2016-NIPS, Chen-2017-ICML, Hochreiter-2001-ICANN, Ravi-2017-ICLR} or learning to initialize the network parameters so that the models can rapidly adapt to novel classes or new tasks~\cite{finn2017model}. Memory-based approaches leverage memory-augmented models (e.g., the hidden activations in a recurrent network or external memory) to retain the learned knowledge~\cite{Kaiser-2018-ICLR, Munkhdalai-2017-ICML, Santoro-2016-ICML}, and associate the learned knowledge with the newly encountered tasks for rapid generalization. Metric-based algorithms aim at learning a feature embedding with proper distance metrics for few-shot~\cite{Snell-2017-NIPS,Sung-2018-CVPR} or one-shot~\cite{Vinyals-2016-NIPS} classification. 

Similar to metric-based meta-learning algorithms, our method also aims at learning a feature embedding. Our method differs from existing meta-learning for visual classification approaches in that we learn a feature embedding from both labeled and unlabeled data with disjoint label sets. Moreover, both the support and query sets share the same label set in most other meta-learning approaches, while the label sets of our meta-training and meta-validation sets are disjoint.

%% file: method.tex
\section{Proposed Method}

\subsection{Algorithmic Overview}

We first describe the setting of our semi-supervised learning task, and define the notations. When matching image pairs in the tasks of image retrieval and person re-ID, we assume that our training set contains a set of $N_L$ labeled images $X_L = \{x_i^L\}_{i=1}^{N_L}$ with the corresponding labels $Y_L = \{y_i^L\}_{i=1}^{N_L}$, and a set of $N_U$ unlabeled images $X_U = \{x_j^U\}_{j=1}^{N_U}$. For the labeled data, each $x_i^L \in \mathbb{R}^{H \times W \times 3}$ and $y_i^L \in \mathbb{R}$ denote the $i^\mathrm{th}$ image and the associated label, respectively. As for $x_j^U \in \mathbb{R}^{H \times W \times 3}$, it is the $j^\mathrm{th}$ unlabeled image in $X_U$. Note that the class number of the labeled set is denoted as $C_L$, while that of the unlabeled set is \emph{unknown}. We assume the label sets of the labeled and unlabeled sets are \emph{disjoint}.

\begin{figure}[t]
  \begin{subfigure}{\linewidth}
    \begin{center}
      \includegraphics[width=\linewidth]{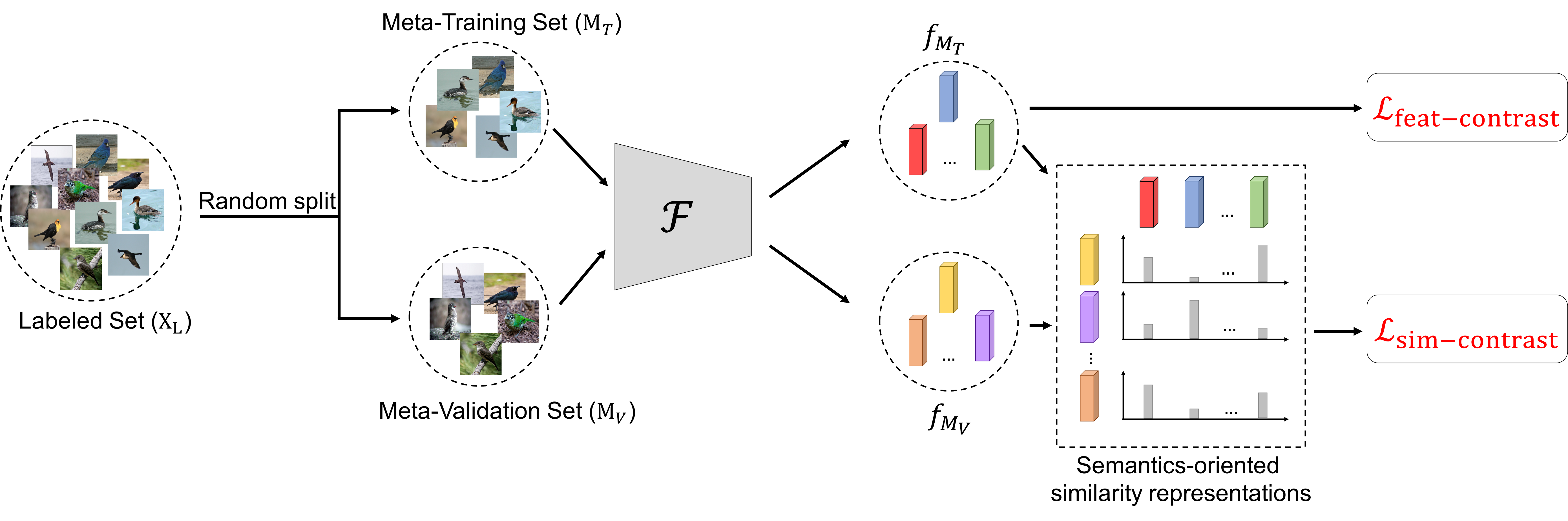}
      \caption{Meta-learning with the labeled training set $X_L$.}
      \label{fig:meta-train}
    \end{center}
  \end{subfigure}
  \begin{subfigure}{\linewidth}
    \begin{center}
      \includegraphics[width=\linewidth]{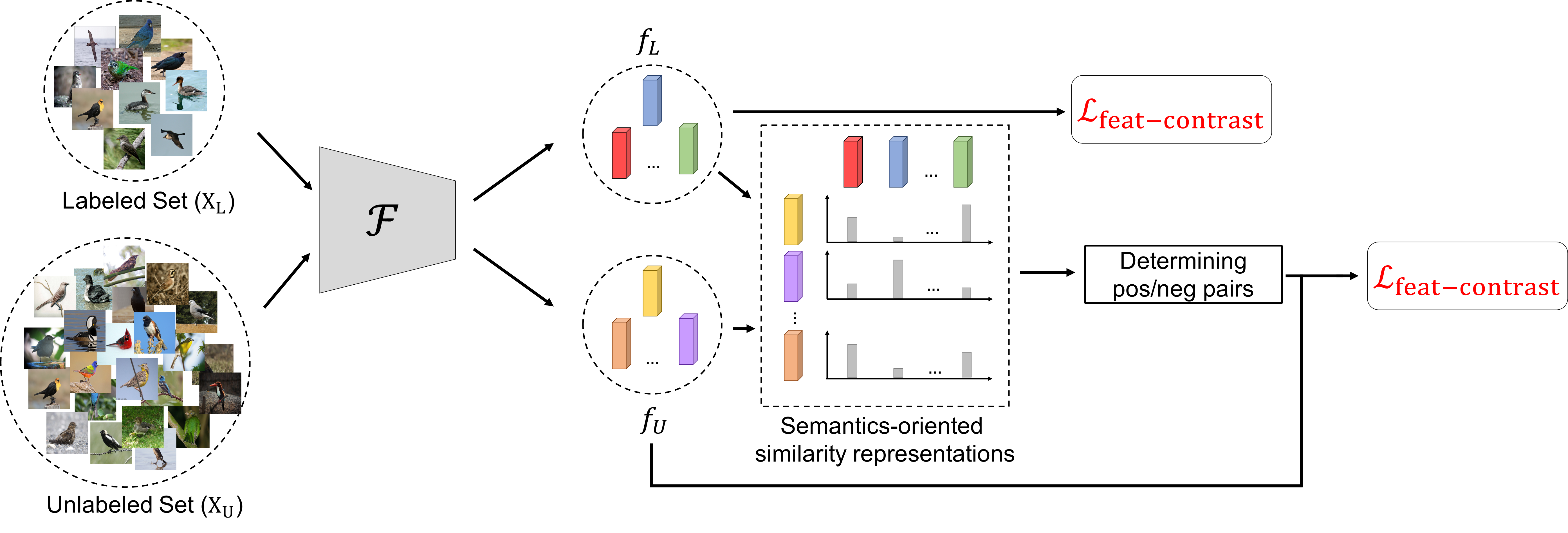}
      \caption{Meta semi-supervised learning on both labeled set $X_L$ and unlabeled set $X_U$.}
      \label{fig:fine-tune}
    \end{center}
  \end{subfigure}
  \caption{\textbf{Overview of the proposed meta-learning framework.} (a) Our model first takes labeled data and learns semantics-oriented similarity representation in a learning-to-learn fashion (i.e., joint learning of $\mathcal{L}_\mathrm{feat-contrast}$ and $\mathcal{L}_\mathrm{sim-contrast}$). (b) The learned concept of matching semantics information allows us to learn from both labeled and unlabeled data, determining positive/negative pairs for computing $\mathcal{L}_\mathrm{feat-contrast}$ for all the training data pairs. Note that the meta-training and meta-validation sets in (a) do \textit{not} share the same labels, \textit{neither} do the labeled and unlabeled training sets in (b).}
  \label{fig:model}
\end{figure}

The goal of this work is to learn a feature embedding model by jointly observing the above labeled and unlabeled sets, with the learned features can be applied for matching images for tasks of retrieval and re-ID. As shown in Figure~\ref{fig:model}, our proposed algorithm comprises two learning phases: 1) meta-learning with labeled training data and 2) meta semi-supervised learning on both labeled and unlabeled training sets. For the first phase (i.e., Figure~\ref{fig:meta-train}), we first partition the labeled set $X_L$ into a meta-training set $M_T = \{x_k^{M_T}\}_{k=1}^{N_{M_T}}$ and a meta-validation set $M_V = \{x_l^{M_V}\}_{l=1}^{N_{M_V}}$, with disjoint labels for $M_T$ and $M_V$ (from $Y_L$). The numbers of images for $M_T$ and $M_V$ are denoted as $N_{M_T}$ and $N_{M_V}$, and the numbers of classes for $M_T$ and $M_V$ are denoted as $C_{M_T}$ and $C_{M_V}$, respectively, summing up as $C_L$ (i.e., $C_L = C_{M_T} + C_{M_V}$). Our model $F$ takes images $x$ from $M_T$ and $M_V$ as inputs, and learns feature representations $f = F(x) \in \mathbb{R}^{d}$ ($d$ is the dimension of $f$) for input images. Our model then derives semantics-oriented similarity representation $s \in \mathbb{R}^{C_{M_T}}$ ($C_{M_T}$ denotes the dimension of $s$) for each image in $M_V$. In the second meta-learning stage (i.e., Figure~\ref{fig:fine-tune}), we utilize the learned concept of semantics-oriented similarity representation to guide the learning of the unlabeled set. The details of each learning phase are elaborated in the following subsections.

As for testing, our model takes a query image as input and extracts its feature $f \in \mathbb{R}^d$, which is applied to match gallery images via nearest neighbor search.

\subsection{Meta-Learning on $X_L$}

Motivated by \cite{yu2019unsupervised}, we exploit the idea of leveraging information from class-wise similarity to guide the learning of the unlabeled data. In our work, we choose to implicitly learn semantics-oriented similarity representation instead of explicit label-specific representation in a learning-to-learn fashion, so that the learned representation can be applied for describing the unlabeled images.

To achieve this, we advance an episodic learning paradigm as applied in existing meta-learning algorithms~\cite{balaji2018metareg,finn2017model}. In each episode, we first divide the labeled set $L$ into a meta-training set $M_T$ and a meta-validation set $M_V$, where the labels of $M_T$ and $M_V$ are \emph{not} overlapped. To learn a feature embedding for matching images, we follow existing methods~\cite{chopra2005learning,hadsell2006dimensionality} and introduce a feature contrastive loss $\mathcal{L}_\mathrm{feat-contrast}$ in the meta-training set $M_T$. That is, given a pair of images $x_i^{M_T}$ and $x_j^{M_T}$ in $M_T$, the feature contrastive loss for $M_T$ is defined as
\begin{equation}
  \mathcal{L}_\mathrm{feat-contrast}(M_T; F) = t \cdot \|f_i^{M_T} - f_j^{M_T}\|  + (1-t) \cdot \max(0, \phi - \|f_i^{M_T} - f_j^{M_T}\|),
\end{equation}
where $t = 1$ if $x_i^{M_T}$ and $x_j^{M_T}$ are of the same label, otherwise $t = 0$, and $\phi > 0$ denotes the margin.

{\flushleft {\bf Semantics-oriented similarity representation $s$.}}
To learn semantics-oriented similarity representation, we first sample a reference image $\hat{x}^{M_T}$ from each class in the meta-training set $M_T$. The sampled reference image for the $k^\mathrm{th}$ class in $M_T$ is denoted as $\hat{x}_k^{M_T}$, and there are $C_{M_T}$ sampled reference images in total. We then extract feature $\hat{f}^{M_T} = F(\hat{x}^{M_T})$ for each reference image $\hat{x}^{M_T}$.

Given an image $x_i^{M_V}$ in the meta-validation set $M_V$, we first extract its feature $f_i^{M_V} = F(x_i^{M_V})$. To learn semantics-oriented similarity representation $s_i^{M_V} \in \mathbb{R}^{C_{M_T}}$ for image $x_i^{M_V}$, we compute the class-wise similarity scores between $f_i^{M_V}$ and all reference features $\hat{f}^{M_T}$ sampled from the meta-training set. The $k^\mathrm{th}$ entry of the semantics-oriented similarity representation $s_i^{M_V}$ is defined as
\begin{equation}
  s_i^{M_V}(k) = \mathrm{sim}(f_i^{M_V}, \hat{f}_k^{M_T}),
\end{equation}
where $\mathrm{sim}(f_i^{M_V}, \hat{f}_k^{M_T})$ denotes the similarity between feature $f_i^{M_V}$ and the sampled reference feature $\hat{f}_k^{M_T}$. We note that we do not limit the similarity measurement in the above equation. For example, we compute the cosine similarity for image retrieval and calculate the $\ell_2$ distance for person re-ID. 

To achieve the learning of semantics-oriented similarity representation, we utilize the ground truth label information from the meta-validation set, and develop a similarity contrastive loss $\mathcal{L}_\mathrm{sim-contrast}$. Specifically, given an image pair $x_i^{M_V}$ and $x_j^{M_V}$ in $M_V$, the associated similarity contrastive loss is defined as
\begin{equation}
  \mathcal{L}_\mathrm{sim-contrast}(M_V; F) = t \cdot \|s_i^{M_V} - s_j^{M_V}\| + (1-t) \cdot \max(0, \phi - \|s_i^{M_V} - s_j^{M_V}\|),
\end{equation}
where $t = 1$ if $x_i^{M_V}$ and $x_j^{M_V}$ are of the same category, otherwise $t = 0$. $\phi > 0$ denotes the margin.

By repeating the above procedure across multiple episodes until the convergence of the meta-validation loss (i.e., the similarity contrastive loss), our model carries out the learning of semantics-oriented similarity representation in a learning-to-learn fashion, without fitting particular class label information. Utilizing such representation allows our model to realize joint learning of labeled and unlabeled data, as discussed next.

\subsection{Meta Semi-Supervised Learning on $X_L$ and $X_U$}

In the semi-supervised setting, learning from labeled data $X_L$ simply follows the standard feature contrastive loss $\mathcal{L}_\mathrm{feat-contrast}(X_L; F)$. To jointly exploit labeled and unlabeled data, we advance the aforementioned meta-training strategy and start from randomly sampling $C_{M_T}$ categories from the labeled set $X_L$. For each sampled class, we then randomly sample one reference image $\hat{x}_k^L$ and extract its feature $\hat{f}_k^L = F(\hat{x}_k^L)$, where $\hat{x}_k^L$ denotes the sampled reference image of the $k^\mathrm{th}$ sampled class. Namely, there are $C_{M_T}$ sampled reference images in total. 

Next, given an image $x_i^{U}$ in the unlabeled set $U$, we extract its feature $f_i^U = F(x_i^U)$, followed by computing the semantics-oriented similarity representation $s_i^U \in \mathbb{R}^{C_{M_T}}$ between $f_i^U$ and all features $\hat{f}^L$ of the above sampled reference images from the labeled set. It is worth repeating that our learning scheme is very different from existing methods~\cite{yu2019unsupervised}, which focus on fitting class-wise similarity scores on the entire labeled set. Instead, we only compute the similarity scores between features of sampled classes. This is the reason why we view our representation to be semantics-oriented instead of class-specific (as~\cite{yu2019unsupervised} does).

Now, we are able to measure the similarity between semantics-oriented similarity representations $s_i^U$ and $s_j^U$, with a threshold $\psi$ to determine whether the corresponding input images $x_i^{U}$ and $x_j^{U}$ are of the same class ($t = 1$) or not ($t = 0$). Namely,
\begin{equation}
  \begin{cases}
    t = 1, \text{ if } \|s_i^U - s_j^U\| < \psi,\\
    t = 0, \text{ otherwise}.
  \end{cases}
  \label{eq:threshold}
\end{equation}

The above process can be viewed as assigning pseudo positive/negative labels for the unlabeled data $X_U$, allowing us to compute the feature contrastive loss $\mathcal{L}_\mathrm{feat-contrast}(X_U; F)$ on any image pair from the unlabeled set.

%% file: experiment.tex
\section{Experiments}

We present quantitative and qualitative results in this section. In all of our experiments, we implement our model using PyTorch and train our model on a single NVIDIA TITAN RTX GPU with $24$ GB memory. The performance of our method can be possibly further improved by applying pre/post-processing methods, attention mechanisms, or re-ranking techniques. However, such techniques are not used in all of our experiments. 

\subsection{Datasets and Evaluation Metrics}

We conduct experiments on four public benchmarks, including the CUB-$200$~\cite{Wah-2011-cub200}, Car$196$~\cite{krause20133d}, Market-$1501$~\cite{zheng2015scalable}, and DukeMTMC-reID~\cite{Ristani_2016_ECCVW} datasets.

{\flushleft {\bf Datasets.}}
For image retrieval, we adopt the CUB-$200$~\cite{Wah-2011-cub200} and Car$196$~\cite{krause20133d} datasets. The CUB-$200$ dataset~\cite{Wah-2011-cub200} is a fine-grained bird dataset containing $11,788$ images of $200$ bird species. Following existing methods~\cite{Song_2016_CVPR,Oh-2017-CVPR,Harwood-2017-ICCV}, we use the first $100$ categories with $5,864$ images for training and the remaining $100$ categories with $5,924$ images for testing. The Car$196$~\cite{krause20133d} dataset is a fine-grained car dataset consisting of $16,189$ images with $196$ car categories. Following \cite{Song_2016_CVPR,Oh-2017-CVPR,Harwood-2017-ICCV}, we use the first $98$ categories with $8,054$ images for training while the remaining $98$ categories with $8,131$ images are used for testing. 

As for person re-ID, we consider the Market-$1501$~\cite{zheng2015scalable} dataset, which contains $32,668$ labeled images of $1,501$ identities captured by $6$ camera views. This dataset is partitioned into a training set of $12,936$ images from $751$ identities, and a test set of $19,732$ images from the other $750$ identities. We also have the DukeMTMC-reID~\cite{Ristani_2016_ECCVW} dataset which is composed of $36,411$ labeled images of $1,404$ identities collected from $8$ camera views. We utilize the benchmarking training/test split, where the training set consists of $16,522$ images of $702$ identities, and the test set contains $19,889$ images of the other $702$ identities.

{\flushleft {\bf Evaluation metrics.}}
Following recent image retrieval methods~\cite{Harwood-2017-ICCV,Song_2016_CVPR}, we use the Recall@K (R@K) metric and the normalized mutual information (NMI)~\cite{Christopher-2008} with cosine similarity for evaluating image retrieval performance. For person re-ID, we adopt the standard single-shot person re-ID setting~\cite{liao2015person} and use the average cumulative match characteristic (CMC) and the mean Average Precision (mAP) with Euclidean distance as similarity measurements.

\subsection{Evaluation of Semi-Supervised Learning Tasks} 

\subsubsection{Image Retrieval} \hfill

{\flushleft {\bf Implementation details and settings.}}
Following \cite{Movshovitz-2017-ICCV, Song_2016_CVPR}, we adopt an ImageNet-pretrained Inception-v1~\cite{Szegedy-2015-CVPR} to serve as the backbone of our model $F$. A fully connected layer with $\ell_2$ normalization is added after the \texttt{pool5} layer to serve as the feature embedding layer. All images are resized to $256 \times 256 \times 3$ in advance. During the first stage of meta-learning, we set the batch sizes of the meta-training and meta-validation sets to $32$ and $64$, respectively. We use the Adam optimizer to train our model for $600$ epochs. The initial learning rate is set to $2 \times 10^{-5}$ and the momentum is set to $0.9$. The learning rate is decreased by a factor of $10$ for every $150$ epochs. The margin $\phi$ is set to $0.3$. As for the meta semi-supervised learning stage, we set the batch size of the labeled set to $32$, the batch size of the unlabeled set to $64$, and the initial learning rate to $1 \times 10 ^{-5}$. Similarly, the learning rate is decayed by a factor of $10$ for every $150$ epochs. We train our model for another $600$ epochs. The similarity threshold $\psi$ is set to $0.01$. We evaluate our method with three different label ratios, i.e., $25\%$, $50\%$, and $75\%$ of the categories are labeled, while the remaining categories are unlabeled.

{\flushleft {\bf Results.}}
We compare our method with existing fully supervised and unsupervised methods. Table~\ref{table:semi-retrieval} reports the results recorded at Recall@$1$, $2$, $4$, and $8$, and NMI on the CUB-$200$~\cite{Wah-2011-cub200} and Car$196$~\cite{krause20133d} datasets. We note that while the results of our method (semi-supervised setting) are not directly comparable to those of fully supervised and unsupervised approaches, their results can be viewed as upper (for fully supervised methods) and lower (for unsupervised approaches) bounds of our results.

The results on both datasets show that our method performs favorably against all competing unsupervised approaches and achieves competitive or even better performance when compared with fully supervised methods.

\input{table/semi-retrieval.tex}

\newpage

\subsubsection{Person re-ID} \hfill

{\flushleft {\bf Implementation details and settings.}}
Following \cite{xin2019semi}, our model $F$ employs an ImageNet-pretrained ResNet-50~\cite{he2016deep}. All images are resized to $256 \times 128 \times 3$ in advance. During the first stage of meta-learning, we set the batch sizes of the meta-training and meta-validation sets to $32$ and $128$, respectively. We use the Adam optimizer to train our model for $600$ epochs. The initial learning rate is set to $2 \times 10^{-3}$, and is decayed by a factor of $10$ for every $150$ epochs. The momentum and the margin $\phi$ are set to $0.9$ and $0.3$, respectively. As for the meta semi-supervised learning stage, we set the batch size of the labeled set to $32$, the batch size of the unlabeled set to $128$, and the initial learning rate to $2 \times 10 ^{-5}$. Similarly, the learning rate is decreased by a factor of $10$ for every $150$ epochs. We train our model for another $600$ epochs. The similarity threshold $\psi$ is set to $0.5$. Also following \cite{xin2019semi}, we evaluate our method with three different label ratios, i.e., $\frac{1}{3}$, $\frac{1}{6}$, and $\frac{1}{12}$ of the person IDs are fully labeled, while the remaining person IDs are unlabeled.

\input{table/semi-reID.tex}

{\flushleft {\bf Results.}}
We compare our method with unsupervised approaches~\cite{liao2015person,zheng2015scalable}, semi-supervised methods~\cite{xin2019semi,xin2019deep}, a fully supervised approach~\cite{zhang2017alignedreid}, and a cross-dataset person re-ID method~\cite{yu2019unsupervised}. Similarly, the results of fully supervised/unsupervised methods can be regarded as the upper/lower bounds of our results. For the cross-dataset person re-ID method~\cite{yu2019unsupervised}, we use their official implementation\footnote{\url{https://github.com/KovenYu/MAR}} with their default hyperparameter settings, and set the labeled set as their source domain and the unlabeled set as their target domain. Table~\ref{table:semi-reID} compares the rank $1$ and mAP scores on the Market-$1501$~\cite{zheng2015scalable} and DukeMTMC-reID~\cite{Ristani_2016_ECCVW} datasets.

From this table, when comparing to semi-supervised learning methods, i.e., MVC~\cite{xin2019semi} and SPMVC~\cite{xin2019deep}, our method consistently outperforms their results by large margins on all three evaluated label ratios of both datasets. When comparing to a fully-supervised method, e.g., AlignedReID~\cite{zhang2017alignedreid}, our method achieves $90\%$ and $85\%$ of their results recorded at rank $1$ on the Market-$1501$~\cite{zheng2015scalable} and DukeMTMC-reID~\cite{Ristani_2016_ECCVW} datasets, respectively, using relatively fewer labeled information, i.e., only $\frac{1}{3}$ of the person IDs are labeled. From these results, we show that under the same experimental setting, our method achieves the state-of-the-art performance, while resulting in comparable results compared to fully supervised approaches.

\input{table/meta-reID.tex}

\subsection{Evaluation of Supervised Learning Tasks} 

\subsubsection{Evaluation of Limited Labeled Data} \hfill

\noindent In addition to evaluating the performance of our semi-supervised learning, we now apply our meta-learning strategy to the labeled training set only and see whether our learning-to-learn strategy would benefit such scenario. 

{\flushleft {\bf Implementation details.}}
All images are resized to $256 \times 128 \times 3$ in advance. We set the batch sizes of the meta-training and meta-validation sets to $32$ and $128$, respectively. We use the Adam optimizer to train our model for $600$ epochs. The initial learning rate is set to $2 \times 10^{-3}$, and is decayed by a factor of $10$ for every $150$ epochs. The momentum and margin $\phi$ are set to $0.9$ and $0.3$, respectively. 

\input{figure/fig_tsne.tex}

\input{table/fully-reID.tex}

{\flushleft {\bf Results.}}
We adopt the Market-$1501$~\cite{zheng2015scalable} and DukeMTMC-reID~\cite{Ristani_2016_ECCVW} datasets for performance evaluations and compare our method with a number of supervised approaches~\cite{martinel2019aggregating,schroff2015facenet,lu2017discriminative,wang2018learning,luo2019bag,zhang2017alignedreid}. Table~\ref{table:meta-reid} presents the experimental results. The results show that our method consistently performs favorably against all competing approaches, demonstrating sufficient re-ID ability can be exhibited by our proposed method even when only limited labeled data are observed.

{\flushleft {\bf Visualization of the learned representations.}}
To demonstrate that our model benefits from learning semantics-oriented similarity representation $s$, we select $20$ person IDs and visualize both semantics-oriented similarity representation $s$ and the learned feature representation $f$ on the Market-$1501$~\cite{zheng2015scalable} test set via t-SNE in Figure~\ref{fig:tsne}, in which we compare our approach with AlignedReID~\cite{zhang2017alignedreid} and its variant method.

We observe that without learning the semantics-oriented similarity representation, AlignedReID~\cite{zhang2017alignedreid} and its variant method cannot separate the representation~$s$ well. Our method, on the other hand, learns semantics-oriented similarity representations from the labeled set in a learning-to-learn fashion. The learned similarity representation $s$ allows our model to guide the learning of the unlabeled set, resulting in a well-separated space for the feature representation $f$.

\subsubsection{Extension to Fully-Supervised Learning Tasks} \hfill

\noindent Finally, to show that our formulation is not limited to semi-supervised learning settings, we apply our learning algorithm to fully-supervised setting on the Market-$1501$~\cite{zheng2015scalable} and DukeMTMC-reID~\cite{Ristani_2016_ECCVW} datasets.

{\flushleft {\bf Results.}}
We initialize our model from AlignedReID~\cite{zhang2017alignedreid} and BoT~\cite{luo2019bag}, respectively, and apply our meta-learning strategy on the entire training set, i.e., there are two variant methods: (1) AlignedReID~\cite{zhang2017alignedreid} + Ours and (2) BoT~\cite{luo2019bag} + Ours. As shown in Table~\ref{table:fully-reID}, our method further improves the performance of AlignedReID~\cite{zhang2017alignedreid} and BoT~\cite{luo2019bag} on both datasets, respectively, comparing favorably against existing fully-supervised learning methods.

\subsection{Limitations and Potential Issues} 

We observe that our method is memory intensive as learning from the unlabeled set requires larger batch size to increase the likelihood of selecting positive image pairs (sampling a negative pair is easier than sampling a positive pair). On the other hand, our learning algorithm is suitable for solving tasks where the categories are visually similar.

%% file: table/semi-retrieval.tex
\begin{table}[!t]
  \small
  \ra{1.3}
  \begin{center}
  \caption{\textbf{Results of semi-supervised image retrieval.} The bold and underlined numbers indicate top two results, respectively.}
  \label{table:semi-retrieval}
  \resizebox{\linewidth}{!} 
  {
  \begin{tabular}{lccccccccccc}
  \toprule
  \multirow{2}{*}{Method} & \multirow{2}{*}{Supervision} & \multicolumn{5}{c}{CUB-200~\cite{Wah-2011-cub200}} & \multicolumn{5}{c}{Car196~\cite{krause20133d}} \\ 
   &  & R@1 & R@2 & R@4 & R@8 & NMI & R@1 & R@2 & R@4 & R@8 & NMI \\ 
  \midrule
  %
  % Contrastive~\cite{chopra2005learning} & \multirow{18}{*}{Supervised} & 27.2 & 36.3 & 49.8 & 62.1 & 47.2 & - & - & - & - & - \\
  %
  % DDML~\cite{lu2017discriminative} & & 31.2 & 41.6 & 54.7 & 67.1 & 47.3 & - & - & - & - & - \\
  %
  % Triplet+DAML~\cite{duan2018deep} & & 37.6 & 49.3 & 61.3 & 74.4 & 51.3 & - & - & - & - & - \\
  % 
%   Triplet hard~\cite{schroff2015facenet} & & 40.6 & 52.3 & 64.2 & 75.0 & 53.4 & 53.2 & 65.4 & 74.3 & 83.6 & 55.7 \\
  %
  % Triplet+HDML~\cite{zheng2019hardness} & & 43.6 & 55.8 & 67.7 & 78.3 & 55.1 & - & - & - & - & - \\
  % 
  % Triplet+~\cite{Harwood-2017-ICCV} & & 45.9 & 57.7 & 69.6 & 79.8 & 58.1 & - & - & - & - & - \\
  %
  %   Proxy NCA~\cite{movshovitz2017no} & & 49.2 & 61.9 & 67.9 & 72.4 & - & 73.2 & 82.4 & 86.4 & 87.8 & - \\
  %
  % N-pair~\cite{sohn2016improved} & & 51.9 & 64.3 & 74.9 & 83.2 & - & 71.1 & 79.7 & 86.5 & 91.6 & - \\
  %
  % DVML~\cite{lin2018deep} & & 52.7 & 65.1 & 75.5 & 84.3 & - & 82.0 & 88.4 & 93.3 & \underline{96.3} & - \\
  %
%   DAML~\cite{duan2018deep} & & 52.7 & 65.4 & 75.5 & 84.3 & - & 75.1 & 83.8 & 89.7 & 93.5 & - \\
  %
  %   Stochastic Mining~\cite{suh2019stochastic} & & 56.0 & 68.3 & 78.2 & 86.3 & - & \underline{83.4} & \underline{89.9} & \underline{93.9} & \textbf{96.5} & -\\
%   %
%   HTL~\cite{ge2018deep} & & \underline{57.1} & \underline{68.8} & \underline{78.7} & \underline{86.5} & - & 81.4 & 88.0 & 92.7 & \underline{95.7} & - \\
  %
%   Multi-Sim~\cite{Wang-CVPR-2019} & & \textbf{65.7} & \textbf{77.0} & \textbf{86.3} & \textbf{91.2} & - & \textbf{84.1} & \textbf{90.4} & \textbf{94.0} & \textbf{96.5} & - \\
%   %
  Triplet~\cite{weinberger2009distance} & \multirow{5}{*}{Supervised} & 35.9 & 47.7 & 59.1 & 70.0 & 49.8 & 45.1 & 57.4 & 69.7 & 79.2 & 52.9 \\
  Lifted~\cite{Song_2016_CVPR} & & 46.9 & 59.8 & 71.2 & 81.5 & 56.4 & 59.9 & 70.4 & 79.6 & 87.0 & \underline{57.8} \\
  Clustering~\cite{Oh-2017-CVPR} & & 48.2 & 61.4 & 71.8 & 81.9 & 59.2 & 58.1 & 70.6 & 80.3 & 87.8 & - \\
  Smart+~\cite{Harwood-2017-ICCV} & & \underline{49.8} & \underline{62.3} & \underline{74.1} & \underline{83.3} & \underline{59.9} & \underline{64.7} & \underline{76.2} & \underline{84.2} & \underline{90.2} & - \\
  Angular~\cite{wang2017deep} & & \textbf{53.6} & \textbf{65.0} & \textbf{75.3} & \textbf{83.7} & \textbf{61.0} & \textbf{71.3} & \textbf{80.7} & \textbf{87.0} & \textbf{91.8} & \textbf{62.4} \\
  \midrule
  %
  %   Rot-Only~\cite{gidaris2018unsupervised} & & 42.5 & 55.8 & \underline{68.6} & \underline{79.4} & 49.1 & 33.3 & 44.6 & 56.4 & 68.5 & 32.7 \\
  %
  %   RPML~\cite{dutta2019probabilistic} & & 43.7 & 56.5 & 68.3 & \underline{79.4} & 53.1 & \textbf{44.6} & \textbf{56.9} & \textbf{68.6} & \textbf{79.5} & \textbf{40.0} \\
  %
  Exemplar~\cite{dosovitskiy2015discriminative} & \multirow{5}{*}{Unsupervised} & 38.2 & 50.3 & 62.8 & 75.0 & 45.0 & 36.5 & 48.1 & 59.2 & 71.0 & 35.4 \\
  NCE~\cite{wu2018unsupervised} & & 39.2 & 51.4 & 63.7 & 75.8 & 45.1 & \underline{37.5} & \underline{48.7} & 59.8 & 71.5 & 35.6 \\
  DeepCluster~\cite{caron2018deep} & & 42.9 & 54.1 & 65.6 & 76.2 & 53.0 & 32.6 & 43.8 & 57.0 & 69.5 & \underline{38.5} \\
  MOM~\cite{iscen2018mining} & & \underline{45.3} & \underline{57.8} & \underline{68.6} & \underline{78.4} & \underline{55.0} & 35.5 & 48.2 & \underline{60.6} & \underline{72.4} & \textbf{38.6} \\
  Instance~\cite{ye2019unsupervised} & & \textbf{46.2} & \textbf{59.0} & \textbf{70.1} & \textbf{80.2} & \textbf{55.4} & \textbf{41.3} & \textbf{52.3} & \textbf{63.6} & \textbf{74.9} & 35.8 \\
  \midrule
  \multirow{3}{*}{Ours} & Semi-supervised ($25\%$) & 48.4 & 60.3 & 71.7 & 81.0 & 55.9 & 54.5 & 66.8 & 77.2 & 85.1 & 48.6 \\
  & Semi-supervised ($50\%$) & \underline{50.5} & \underline{61.1} & \underline{72.3} & \underline{82.9} & \underline{57.6} & \underline{62.2} & \underline{73.8} & \underline{83.0} & \underline{89.4} & \underline{55.0} \\
  & Semi-supervised ($75\%$) & \textbf{51.0} & \textbf{62.3} & \textbf{73.4} & \textbf{83.0} & \textbf{59.3} & \textbf{65.9} & \textbf{76.6} & \textbf{84.4} & \textbf{90.1} & \textbf{57.7} \\
  \bottomrule
  \end{tabular}
  }
  \end{center}
\end{table}

%\begin{table}[t]
%  \small
%  \ra{1.3}
%  \begin{center}
%  \caption{\textbf{Results of semi-supervised image retrieval.} We present the results of image retrieval on the CUB-$200$~\cite{Wah-2011-cub200} dataset. The bold and underlined numbers indicate top two results, respectively.}
%  %
%  \label{table:semi-retrieval}
%  \resizebox{\linewidth}{!} 
%  {
%  \begin{tabular}{c|lcccccc}
%  %
%  \toprule
%  %
%  & Method & Label ratio & R@1 & R@2 & R@4 & R@8 & NMI \\ 
%  %
%  \midrule
%  %
%  \multirow{4}{*}{\rotatebox{90}{Sup.}} & Lifted~\cite{Song_2016_CVPR} & \multirow{4}{*}{$100\%$} & 43.6 & 56.6 & 68.6 & 79.6 & 56.5 \\
%  %
%  & Clustering~\cite{Oh-2017-CVPR} & & 48.2 & 61.4 & 71.8 & 81.9 & 59.2 \\
%  %
%  & Triplet+~\cite{Harwood-2017-ICCV} & & 45.9 & 57.7 & 69.6 & 79.8 & 58.1 \\
%  %
%  & Smart+~\cite{Harwood-2017-ICCV} &  & 49.8 & \underline{62.3} & 74.1 & 83.3 & \textbf{59.9} \\
%  %
%  \midrule
%  %
%  \multirow{3}{*}{\rotatebox{90}{Semi-sup.}} & \multirow{3}{*}{Ours} & $25\%$ & 48.4 & 60.3 & 71.7 & 81.0 & 55.9 \\
%  %
%  &  & $50\%$ & \underline{50.5} & \textbf{63.0} & 74.0 & 83.1 & 57.6 \\
%  %
%  &  & $75\%$ & \textbf{51.0} & \underline{62.3} & 73.4 & 83.0 & \underline{59.3} \\
%  %
%  \bottomrule
%  %
%  \end{tabular}
%  }
%  \end{center}
%\end{table}

%% file: table/semi-reID.tex
\begin{table}[!t]
  \small
  \ra{1.3}
  \begin{center}
  \caption{\textbf{Results of semi-supervised person re-ID.} The bold numbers indicate the best results.}
  \label{table:semi-reID}
  \resizebox{\linewidth}{!} 
  {
  \begin{tabular}{lcccccc}
  \toprule
  \multirow{2}{*}{Method} & \multirow{2}{*}{Supervision} & \multirow{2}{*}{Backbone} & \multicolumn{2}{c}{Market-$1501$~\cite{zheng2015scalable}} & \multicolumn{2}{c}{DukeMTMC-reID~\cite{Ristani_2016_ECCVW}} \\
  & & & Rank 1 & mAP & Rank 1 & mAP \\ 
  \midrule
  LOMO~\cite{liao2015person} & \multirow{2}{*}{Unsupervised} & - & 27.2 & 8.0 & 12.3 & 4.8 \\
  BOW~\cite{zheng2015scalable} & & - & 35.8 & 14.8 & 17.1 & 8.3 \\
  \midrule
  AlignedReID~\cite{zhang2017alignedreid} & Supervised & ResNet-50 & 89.2 & 72.8 & 79.3 & 65.6 \\
  \midrule
  MVC~\cite{xin2019semi} & \multirow{2}{*}{Semi-supervised ($\frac{1}{12}$)} & ResNet-50 & \underline{46.6} & - & \underline{34.8} & - \\
  %
%   MAR~\cite{yu2019unsupervised} & & ResNet-50 & 34.2 & \underline{18.8} & - & - \\
  %
  Ours & & ResNet-50 & \textbf{56.7} & \textbf{32.4} & \textbf{44.9} & \textbf{24.4} \\
  \midrule
  MVC~\cite{xin2019semi} & \multirow{2}{*}{Semi-supervised ($\frac{1}{6}$)} & ResNet-50 & \underline{60.0} & - & \underline{43.8} & - \\
  %
%   MAR~\cite{yu2019unsupervised} & & ResNet-50 & 44.8 & \underline{24.5} & - & - \\
  %
  Ours & & ResNet-50 & \textbf{70.8} & \textbf{46.4} & \textbf{56.6} & \textbf{33.6} \\
  \midrule
  MVC~\cite{xin2019semi} & \multirow{4}{*}{Semi-supervised ($\frac{1}{3}$)} & ResNet-50 & \underline{72.2} & 48.7 & 52.9 & 33.6 \\
  SPMVC~\cite{xin2019deep} & & ResNet-50 & 71.5 & \underline{53.2} & \underline{58.5} & \underline{37.4} \\
  MAR~\cite{yu2019unsupervised} & & ResNet-50 & 69.9 & 46.4 & - & - \\
  Ours & & ResNet-50 & \textbf{80.3} & \textbf{58.7} & \textbf{67.5} & \textbf{46.3} \\
  \bottomrule
  \end{tabular}
  }
  \end{center}
\end{table}

%% file: table/meta-reID.tex
\begin{table}[!t]
  \small
  \ra{1.3}
  \begin{center}
  \caption{\textbf{Results of fully-supervised person re-ID with limited training data.} The bold and underlined numbers indicate the top two results, respectively.}
  \label{table:meta-reid}
  \resizebox{\linewidth}{!} 
  {
  \begin{tabular}{l|cccc|cccc|cccc}
  \toprule
  \multirow{3}{*}{Method} & \multicolumn{12}{c}{Market-1501~\cite{zheng2015scalable}} \\
  & \multicolumn{4}{c|}{$\frac{1}{3}$ of the IDs are available} & \multicolumn{4}{c|}{$\frac{1}{6}$ of the  IDs are available} & \multicolumn{4}{c}{$\frac{1}{12}$ of the IDs are available}\\
  & Rank 1 & Rank 5 & Rank 10 & mAP & Rank 1 & Rank 5 & Rank 10 & mAP & Rank 1 & Rank 5 & Rank 10 & mAP \\
  \midrule
  DDML~\cite{lu2017discriminative} & 72.1 & 86.9 & 91.4 & 45.5 & 62.3 & 81.1 & 86.5 & 35.4 & 49.6 & 70.1 & 78.0 & 24.8 \\
  Triplet hard~\cite{schroff2015facenet} & 72.6 & 87.5 & 92.0 & 49.9 & 61.4 & 81.2 & 87.0 & \underline{38.2} & 47.5 & 69.2 & 77.7 & 25.6 \\
  Triplet+HDML~\cite{zheng2019hardness} & 73.2 & 88.6 & 92.3 & 48.0 & 62.0 & 80.7 & 86.9 & 35.3 & 48.0 & 70.0 & 78.9 & 25.2 \\
  %
  %AlignedReID w/o $\mathcal{L}_\mathrm{cls}$~\cite{zhang2017alignedreid} & 70.4 & 87.1 & 91.3 & 48.2 & 59.5 & 80.0 & 86.4 & 37.3 & 46.9 & 69.0 & 77.8 & 25.7 \\
  %
  AlignedReID~\cite{zhang2017alignedreid} & 73.3 & 88.1 & 92.1 & 47.7 & 62.2 & 81.3 & 87.0 & 36.1 & 49.7 & 71.4 & 78.8 & 25.8 \\
  MGN~\cite{wang2018learning} & 74.1 & 88.2 & 92.1 & 50.8 & 62.3 & 81.4 & 86.7 & 38.1 & \underline{50.6} & 71.6 & 79.8 & 27.4
\\
  BoT~\cite{luo2019bag} & 74.8 & \underline{89.7} & \underline{93.5} & 51.8 & 60.6 & 80.3 & 86.5 & 36.7 & 47.1 & 69.0 & 77.8 & 24.2 \\
  PyrNet~\cite{martinel2019aggregating} & \underline{74.9} & \underline{89.7} & 92.7 & \underline{52.1} & \underline{63.2} & \underline{81.8} & \underline{87.2} & 37.8 & 50.2 & \underline{71.7} & \underline{79.9} & \underline{28.4} \\
  Ours & \textbf{77.0} & \textbf{90.8} & \textbf{93.9} & \textbf{54.0} & \textbf{66.0} & \textbf{85.2} & \textbf{90.3} & \textbf{41.2} & \textbf{53.4} & \textbf{74.9} & \textbf{82.2} & \textbf{29.2} \\ 
  \toprule
  & \multicolumn{12}{c}{DukeMTMC-reID~\cite{Ristani_2016_ECCVW}} \\
  \midrule
  PyrNet~\cite{martinel2019aggregating} & 59.7 & 75.8 & 80.5 & 40.6 & 51.6 & 68.0 & 73.5 & 31.9 & 39.8 & 56.6 & 63.5 & \underline{21.2} \\
  Triplet hard~\cite{schroff2015facenet} & 60.6 & 76.5 & 82.3 & 40.1 & 51.8 & \underline{69.1} & 75.5 & 30.1 & 40.0 & \underline{58.2} & \underline{65.4} & 20.3 \\
  DDML~\cite{lu2017discriminative} & 60.6 & 75.0 & 79.1 & 36.7 & 51.5 & 67.3 & 73.2 & 29.6 & 40.1 & 57.4 & 64.3 & 20.1 \\
  MGN~\cite{wang2018learning} & 60.6 & 75.1 & 80.3 & \underline{41.2} & 51.4 & 66.3 & 72.7 & \underline{31.9} & 39.7 & 56.3 & 63.4 & 20.8 \\
  BoT~\cite{luo2019bag} & 61.9 & \underline{78.5} & \underline{83.8} & 40.5 & \underline{52.4} & 68.9 & \underline{75.9} & 31.8 & \underline{40.7} & 57.6 & 64.3 & 21.1 \\
  %
  %AlignedReID w/o $\mathcal{L}_\mathrm{cls}$~\cite{zhang2017alignedreid} & 60.6 & 76.5 & 82.3 & 40.1 & 49.4 & 66.4 & 73.2 & 28.7 & 37.8 & 55.0 & 61.8 & 19.4 \\
  %
  AlignedReID~\cite{zhang2017alignedreid} & \underline{62.5} & 77.1 & 82.2 & 40.3 & 51.3 & 68.2 & 75.6 & 29.6 & 40.5 & 58.1 & \underline{65.4} & 20.5 \\
  Ours & \textbf{64.9} & \textbf{80.9} & \textbf{85.6} & \textbf{44.8} & \textbf{54.0} & \textbf{70.9} & \textbf{76.7} & \textbf{32.1} & \textbf{42.6} & \textbf{60.5} & \textbf{67.3} & \textbf{22.2} \\ 
  \bottomrule
  \end{tabular}
  }
  \end{center}
\end{table}

%% file: figure/fig_tsne.tex
\begin{figure}[!t]
  \begin{center}
  \mpage{0.01}{\raisebox{12pt}{\rotatebox{90}{Representation $s$}}} 
  \mpage{0.3}{\includegraphics[width=0.9\linewidth]{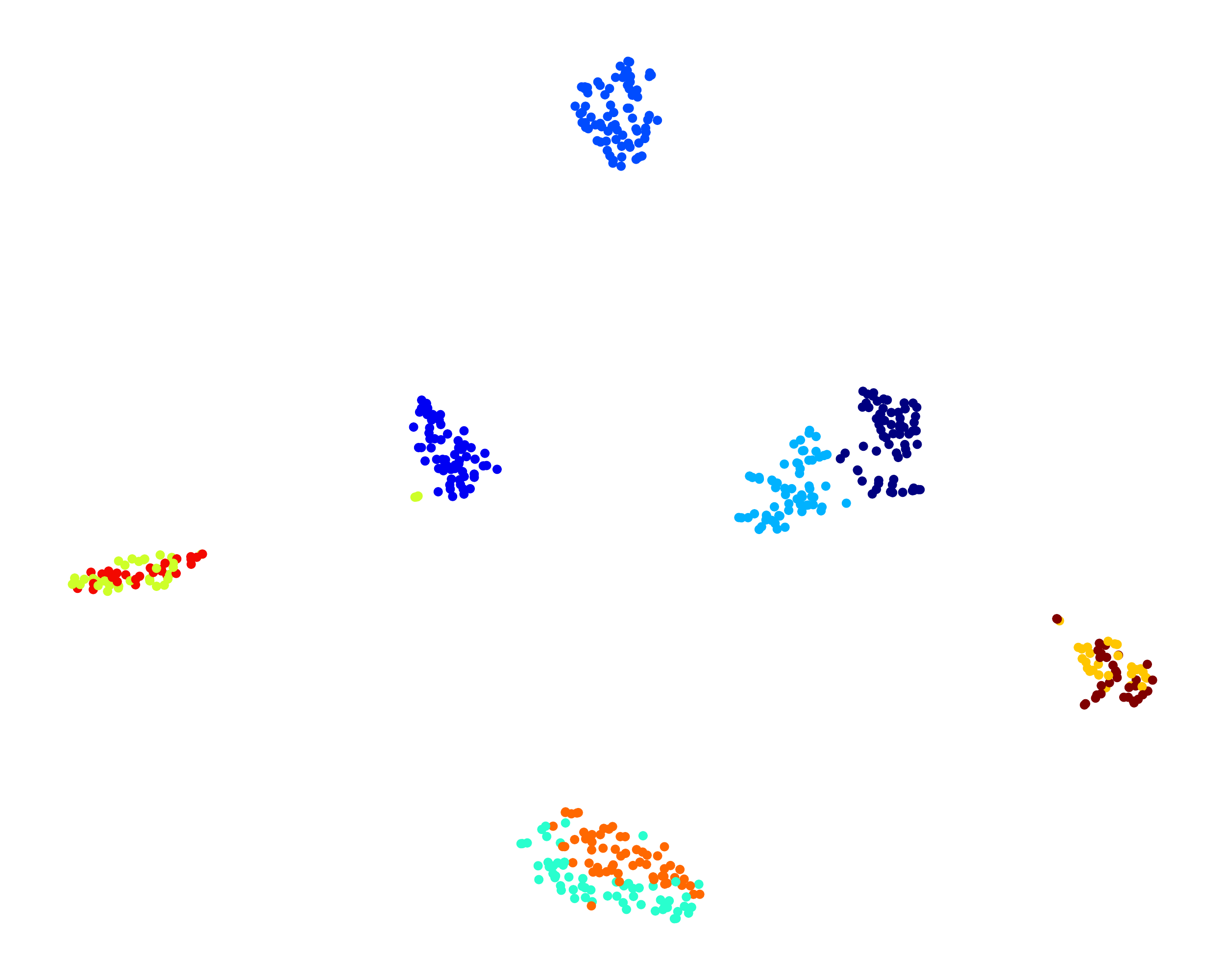}} \hfill
  \mpage{0.3}{\includegraphics[width=0.9\linewidth]{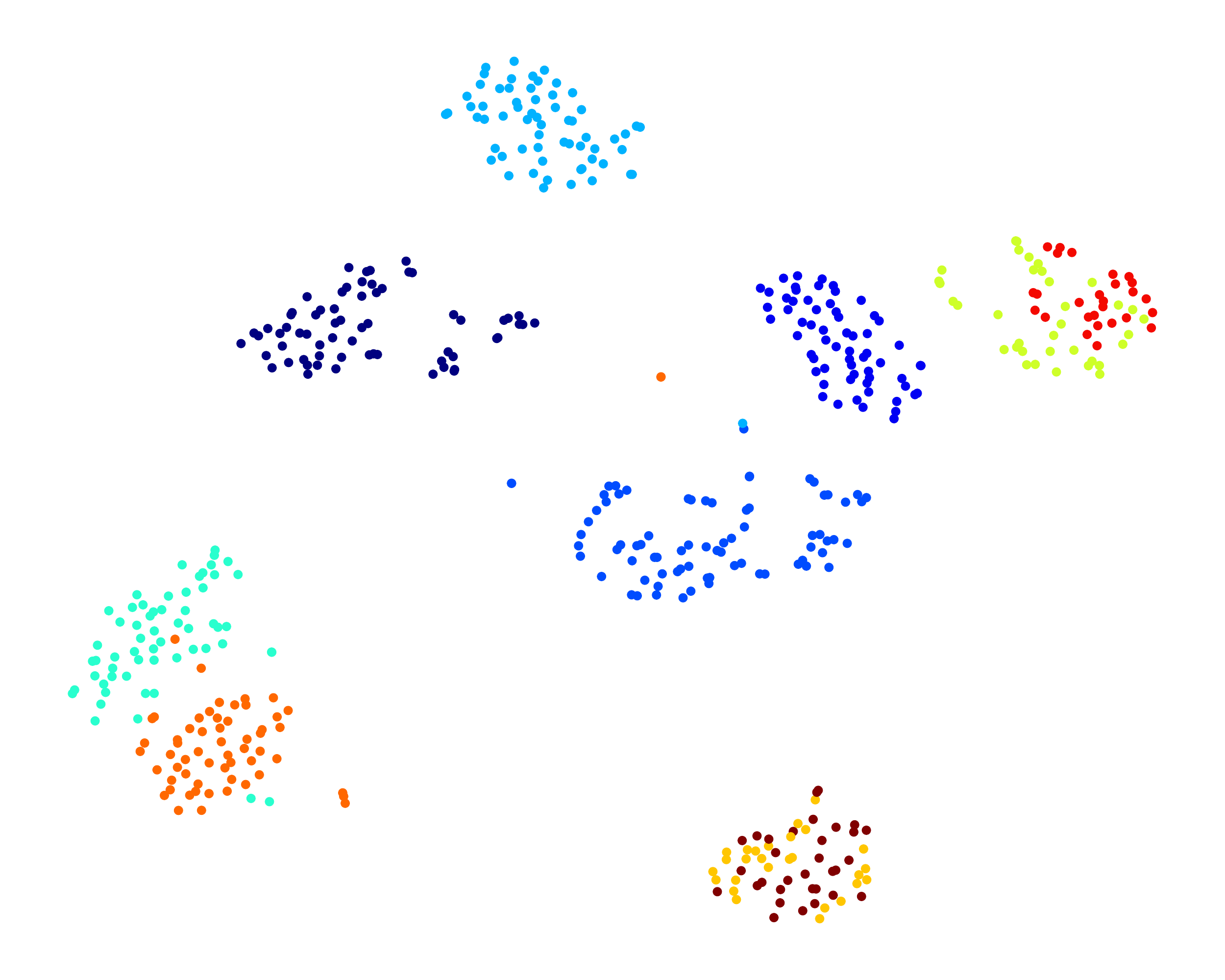}} \hfill
  \mpage{0.3}{\includegraphics[width=0.9\linewidth]{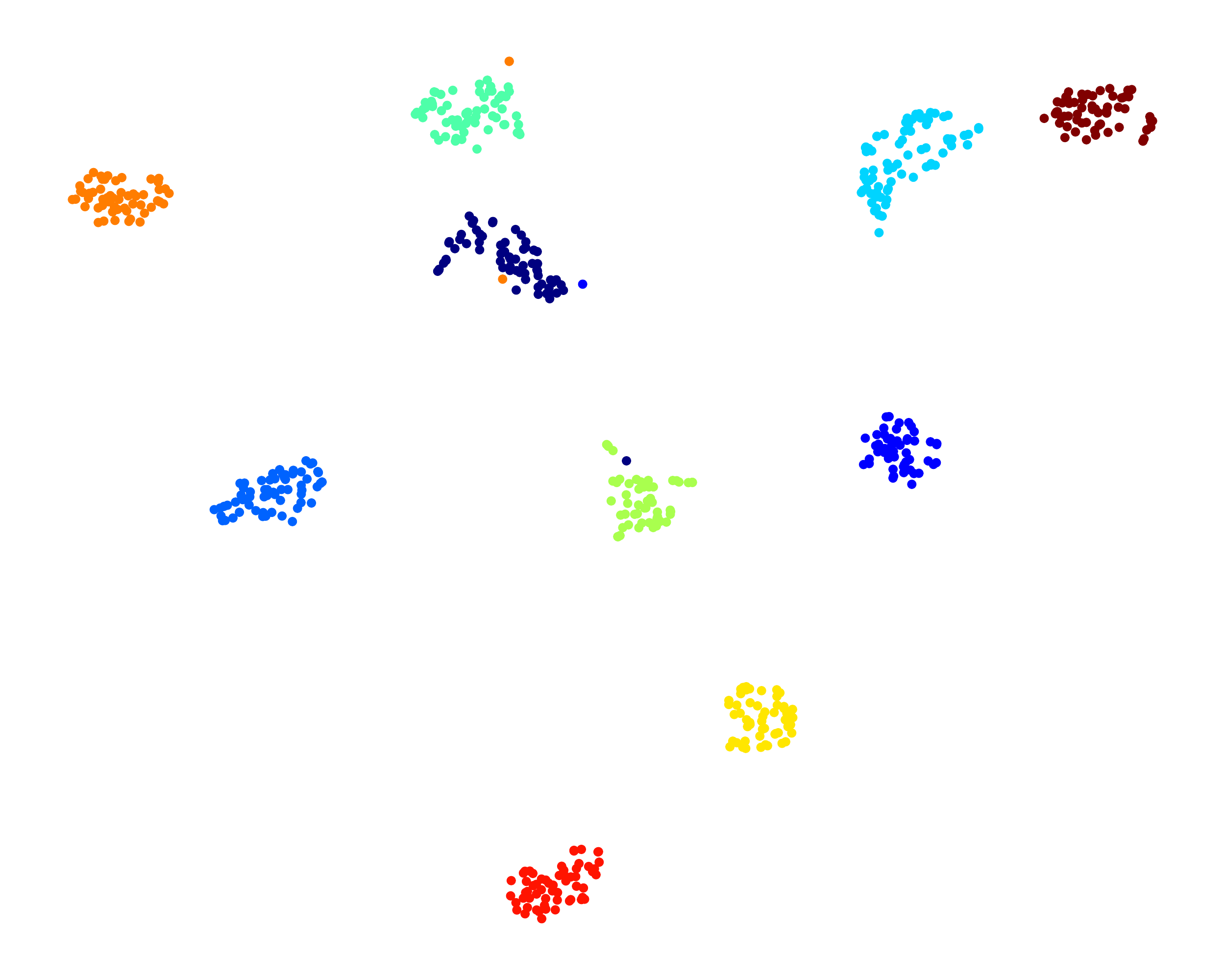}} \\
  \mpage{0.01}{\raisebox{12pt}{\rotatebox{90}{Representation $f$}}} 
  \mpage{0.3}{\includegraphics[width=0.9\linewidth]{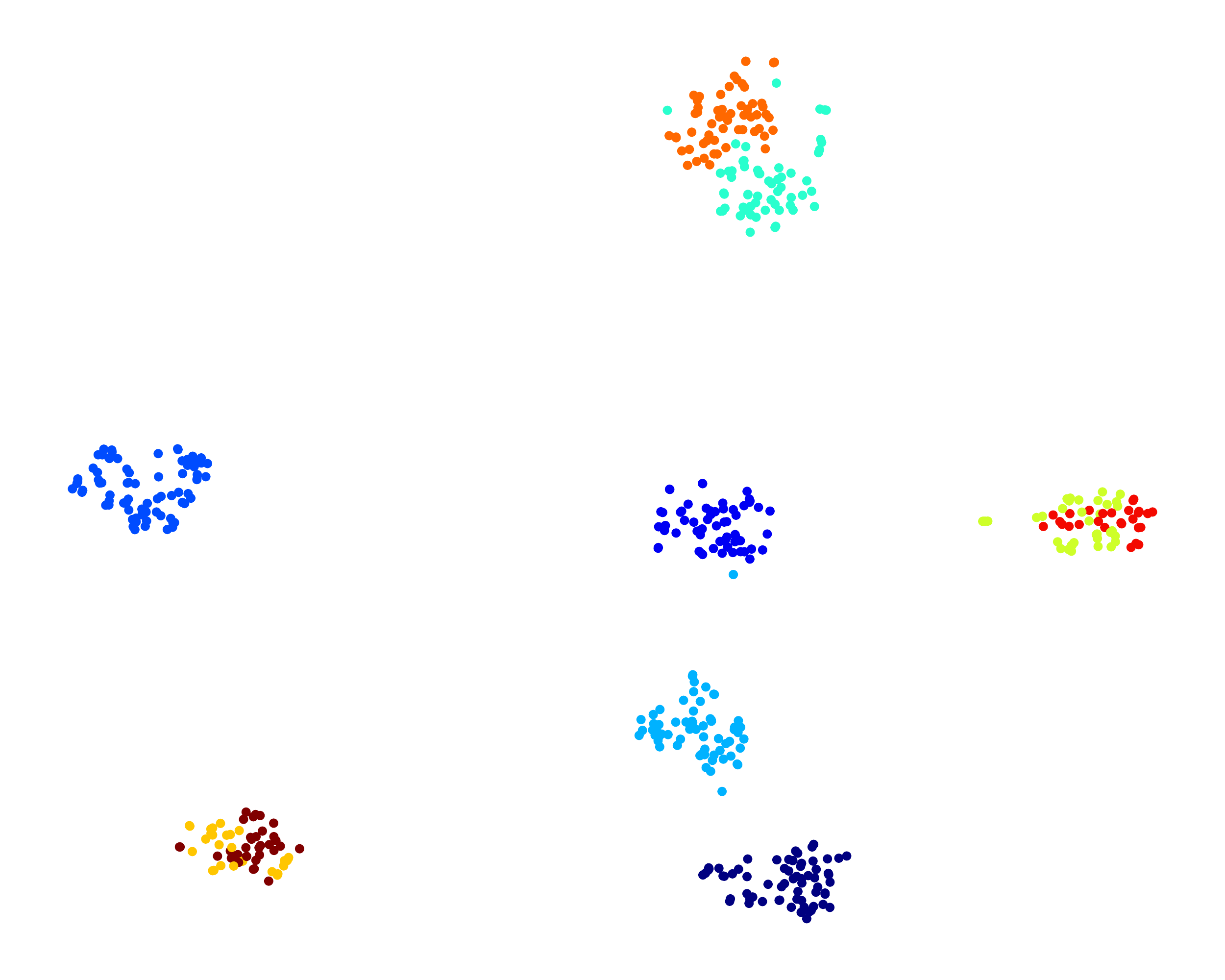}} \hfill
  \mpage{0.3}{\includegraphics[width=0.9\linewidth]{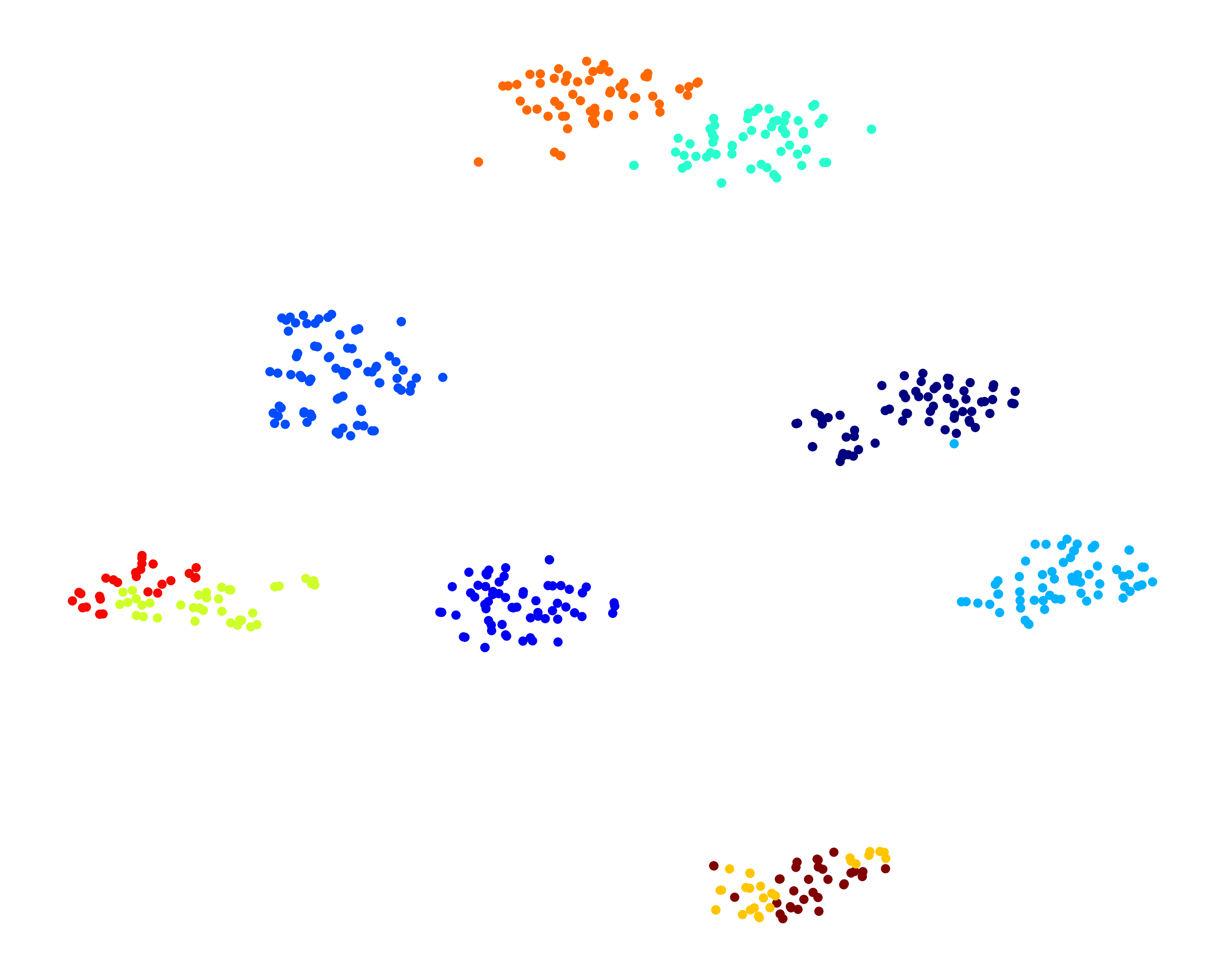}} \hfill
  \mpage{0.3}{\includegraphics[width=0.9\linewidth]{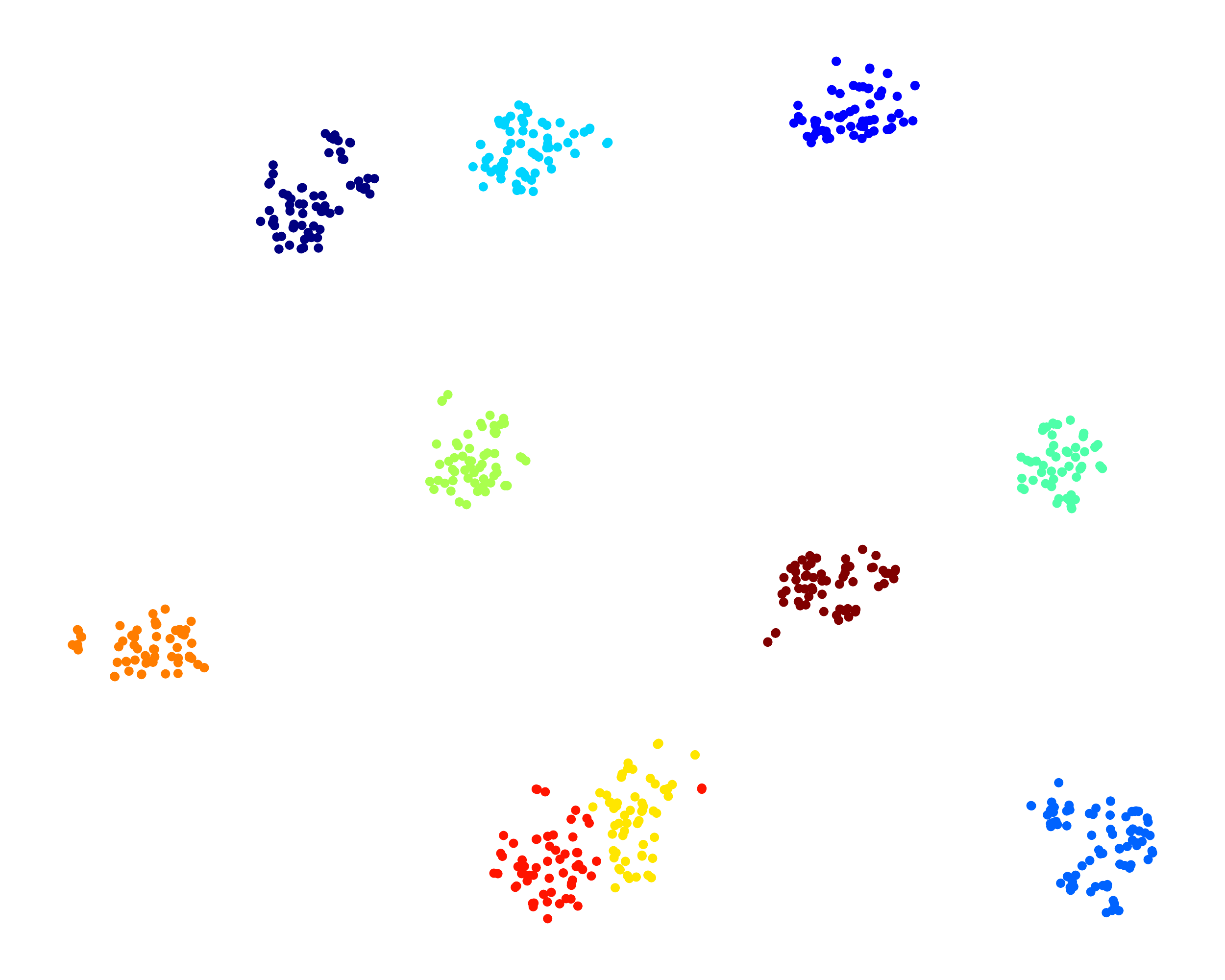}} \\
  \mpage{0.01}{} 
  \mpage{0.3}{AlignedReID~\cite{zhang2017alignedreid} \\ ($\mathcal{L}_\mathrm{feat-contrast}$ only)} \hfill
  \mpage{0.3}{AlignedReID~\cite{zhang2017alignedreid} \\ ($\mathcal{L}_\mathrm{cls}$ and $\mathcal{L}_\mathrm{feat-contrast}$)} \hfill
  \mpage{0.3}{Ours} \\
  \caption{\textbf{Visual comparisons of the learned representations on Market-1501.} (\emph{Top row}) visualizes the semantics-oriented similarity representation. (\emph{Bottom row}) visualizes the feature representation $f$. Note that selected samples of 20 identities are illustrated, each in a specific color. Comparing to AlignedReID, our model only learns semantics similarity and achieves comparable/improved performances.}
  \label{fig:tsne}
  \end{center}
\end{figure}

%% file: table/fully-reID.tex
\begin{table}[!t]
  \small
  \ra{1.3}
  \begin{center}
  \caption{\textbf{Results of fully-supervised person re-ID.} The bold and underlined numbers indicate the top two results, respectively.}
  \label{table:fully-reID}
  %\resizebox{\linewidth}{!} 
  {
  \begin{tabular}{lcccccc}
  \toprule
  \multirow{2}{*}{Method} & \multicolumn{3}{c}{Market-$1501$~\cite{zheng2015scalable}} & \multicolumn{3}{c}{DukeMTMC-reID~\cite{Ristani_2016_ECCVW}} \\ 
  & Rank 1 & Rank 5 & mAP & Rank 1 & Rank 5 & mAP\\ 
  \midrule
  %
%   SpindleNet~\cite{zhao2017spindle} & 76.9 & 91.5 & - & - & - & - \\
%   %
%   HydraPlus-Net~\cite{liu2017hydraplus} & 76.9 & 91.3 & - & - & - & - \\
  %
  Part-Aligned~\cite{zhao2017deeply} & 81.0 & 92.0 & 63.4 & - & - & - \\
  %
%   SVDNet~\cite{Sun-2017-CVPR} & 82.3 & 92.3 & 62.1 & 76.7 & 86.4 & 56.8 \\
  %
  PAN~\cite{Zheng-2018-TCSVT} & 82.8 & 93.5 & 63.4 & 71.6 & 83.9 & 51.5 \\
  MGCAM~\cite{song2018mask} & 83.8 & - & 74.3 & - & - & - \\
  TriNet~\cite{Hermans-2017-arXiv} & 84.9 & 94.2 & 69.1 & - & - & - \\
  JLML~\cite{Li-2017-IJCAI} & 85.1 & - & 65.5 & - & - & - \\
  PoseTransfer~\cite{Zhong-2018-CVPR} & 87.7 & - & 68.9 & 78.5 & - & 56.9 \\
  PSE~\cite{saquib2018pose} & 87.7 & 94.5 & 69.0 & 79.8 & 89.7 & 62.0 \\
  CamStyle~\cite{Zhong-2018-CVPR} & 88.1 & - & 68.7 & 75.3 & - & 53.5 \\
  DPFL~\cite{Cheng-2016-CVPR} & 88.9 & 92.3 & 73.1 & 79.2 & - & 60.6 \\
  AlignedReID~\cite{zhang2017alignedreid} & 89.2 & 95.9 & 72.8 & 79.3 & 89.7 & 65.6 \\
  DML~\cite{Zhang-2018-CVPR} & 89.3 & - & 70.5 & - & - & - \\
  DKP~\cite{shen2018end} & 90.1 & 96.7 & 75.3 & 80.3 & 89.5 & 63.2 \\
  DuATM~\cite{si2018dual} & 91.4 & 97.1 & 76.6 & 81.8 & 90.2 & 68.6 \\
  RDR~\cite{almazan2018re} & 92.2 & 97.9 & 81.2 & 85.2 & \textbf{93.9} & 72.8 \\
  %
%   PyrNet~\cite{martinel2019aggregating} & 93.6 & 98.2 & 81.7 & \textbf{87.1} & \textbf{94.1} & 74.0 \\
  %
  SPReID~\cite{song2018mask} & 93.7 & 97.6 & 83.4 & 85.9 & 92.9 & 73.3 \\
  BoT~\cite{luo2019bag} & \underline{94.5} & \underline{98.2} & \underline{85.9} & \underline{86.4} & \underline{93.6} & \underline{76.4} \\
  \midrule
  AlignedReID~\cite{zhang2017alignedreid} + Ours & 91.1 & 96.3 & 78.1 & 81.7 & 91.0 & 67.7 \\
  BoT~\cite{luo2019bag} + Ours & \textbf{94.8} & \textbf{98.3} & \textbf{86.1} & \textbf{86.6} & \textbf{93.9} & \textbf{76.8} \\
  \bottomrule
  \end{tabular}
  }
  \end{center}
\end{table}

%% file: conclusion.tex
\section{Conclusions}

We presented a meta-learning algorithm for semi-supervised learning with applications to image retrieval and person re-ID. We consider the training schemes in which labeled and unlabeled data share non-overlapping categories. Our core technical novelty lies in learning semantics-oriented similarity representation from the labeled set in a learning-to-learn fashion, which can be applied to semi-supervised settings without knowing the number of classes of the unlabeled data in advance. Our experiments confirmed that our method performs favorably against state-of-the-art image retrieval and person re-ID approaches in semi-supervised settings. We also verified that our algorithm can be applied to supervised settings for improved performance, which further exhibits the effectiveness and applicability of our learning algorithm.